\documentclass{article}
\usepackage{INTERSPEECH2021,amsmath,graphicx,caption,makecell,multirow}

\title{OVERCOMING DOMAIN MISMATCH IN LOW RESOURCE SEQUENCE-TO-SEQUENCE ASR MODELS USING HYBRID GENERATED PSEUDOTRANSCRIPTS}
\name{Chak-Fai Li, Francis Keith, William Hartmann, Matthew Snover, Owen Kimball}
\address{Raytheon BBN Technologies, Cambridge MA, USA}
\email{\{chak.fai.li, francis.keith, william.hartmann, matt.snover, owen.kimball\}@raytheon.com}
\begin{document}
\ninept
\maketitle
\begin{abstract}
Sequence-to-sequence (seq2seq) models are competitive with hybrid models for automatic speech recognition (ASR) tasks when large amounts of training data are available.
However, data sparsity and domain adaptation are more problematic for seq2seq models than their hybrid counterparts.
We examine corpora of five languages from the IARPA MATERIAL program where the transcribed data is conversational telephone speech (CTS) and evaluation data is broadcast news (BN).
We show that there is a sizable initial gap in such a data condition between hybrid and seq2seq models, and the hybrid model is able to further improve through the use of additional language model (LM) data.
We use an additional set of untranscribed data primarily in the BN domain for semisupervised training.
In semisupervised training, a seed model trained on transcribed data generates hypothesized transcripts for unlabeled domain-matched data for further training.
By using a hybrid model with an expanded language model for pseudotranscription, we are able to improve our seq2seq model from an average word error rate (WER) of 66.7\% across all five languages to 29.0\% WER.
While this puts the seq2seq model at a competitive operating point, hybrid models are still able to use additional LM data to maintain an advantage.
\end{abstract}
\noindent\textbf{Index Terms}: Speech recognition, sequence-to-sequence, semisupervised learning
\section{Introduction}
\label{sec:intro}

Recently, automatic speech recognition (ASR) research has seen a dramatic increase in the focus on sequence-to-sequence (seq2seq) models \cite{ibm:swbd300, aachen:e2e}. 
These seq2seq models consist of a single neural network that is able to behave as both the acoustic and language model that would be used in a hybrid DNN-HMM system. 
Additionally, these seq2seq models require no alignments, allowing for a much simpler training procedure. 
While there has been a significant amount of research that shows these models can surpass hybrid systems with large amounts of training data \cite{ibm:swbd300, google:conformer, google:specaug}, seq2seq models are less effective in cases where transcribed data is more limited \cite{ibm:e2ekws}. 

As examples of some of the challenges of adapting seq2seq models to low resource conditions, the WER of seq2seq systems on BABEL languages in \cite{cho2018multilingual, wang2020improving} are 10-20\% absolute worse than competitive hybrid models in \cite{karafidt2018analysis}, and the WER of seq2seq systems on low-resource Indian languages in \cite{shetty2020improving} is 10\% absolute worse than the hybrid baseline in \cite{srivastava2018interspeech}. 
All of \cite{cho2018multilingual, wang2020improving, shetty2020improving} contain compelling and valuable work on low resource learning with seq2seq models, but this comparison highlights the difficulty in using seq2seq models in such conditions compared to hybrid models.

In this paper, we explore some of the challenges of applying seq2seq models on low resource corpora. 
Our particular focus is to explore the effects of a domain mismatch between the training and evaluation data.
%
We examine five languages from the IARPA MATERIAL program\footnote{https://www.iarpa.gov/index.php/research-programs/material}.
The amount of transcribed data for each language is smaller than most corpora used in seq2seq research, and is comprised entirely of transcribed conversational telephone speech (CTS).
The evaluation sets we use for these languages are broadcast news (BN) data, presenting a significant domain mismatch. 
We train seq2seq and hybrid models on the transcribed data and examine some of the differences from these data conditions. 
We show that the gap is quite large when there is a domain mismatch, and attempt to overcome this.
Language models have been shown to be important for low resource seq2seq models \cite{inaguma2019transfer}, so we explore augmenting the language models with various amounts of data. 

For each language, we have a set of untranscribed data that primarily consists of BN data that matches the target domain.
The focal point of this work is using this untranscribed data for semisupervised training \cite{but:semisup} to overcome domain mismatch.
We first use the relatively simple approach of self-training \cite{bbn:semisup, ms:semisup}.
In self-training, transcribed data is used to train an initial seed model, which is used for decoding the untranscribed data to generate \textit{pseudotranscripts}.
These psuedotranscripts are treated as true transcripts and combined with the original data for retraining the acoustic models.
Previous work has demonstrated the effectiveness of semisupervised training for domain adaptation \cite{wotherspoon2021improved}.

However, a strong pseudotranscription seed model is very important for semisupervised training, especially in the case of seq2seq models \cite{Kahn_2020}.
While work has been done on using pseudotranscripts in seq2seq models \cite{Kahn_2020, chen2020semisupervised, park2020improved}, the baseline seq2seq model is typically able to generate reasonable pseudotranscripts. 
We show that this is not the case with models in our data condition---largely due to domain mismatch---and the standard self-learning approach provides only a minimal gain on seq2seq models.
We present a simple alteration to the standard self-learning paradigm that uses pseudotranscripts generated from a well-trained hybrid system for training a seq2seq model.
This semisupervised technique produces a seq2seq model that is competitive with hybrid systems in the mismatched domain, while hybrid semisupervised models are able to maintain an advantage due to their ability to make better use of LM data.

The rest of the paper is organized as follows. 
In Section~\ref{sec:data}, we describe all the data used in our experiments. 
In Section~\ref{sec:models}, we describe the seq2seq and hybrid models we use, along with our semisupervised approach.
Section~\ref{sec:results} goes into more detail on the experiments and provides analysis of the results.
We conclude with some closing thoughts and ideas for future work in Section~\ref{sec:conclusion}.
\section{Data}
\label{sec:data}
\begin{table}[t]
  \captionsetup{justification=centering}
  \caption{\label{tab:acdata} {\it Audio data (hours) for the supervised (CTS only), unsupervised (CTS and BN), and evaluation sets (BN only)}}
  \centerline{
    \begin{tabular}{| l | c | c | c | c | c |}
    \hline
    audio set & Sw. & Li. & Bu. & Ta. & So. \\
    \hline \hline
    sup. CTS & 68.3 & 66.4 & 41.1 & 127.9 & 47.7 \\
    unsup. CTS & 57.6 & 11.4 & 33.3 & 48.4 & 26.5 \\
    unsup. BN & 149.0 & 160.3 & 149.9 & 153.8 & 167.8 \\
    eval. BN & 5.3 & 9.6 & 14.0 & 8.7 & 9.5 \\
    \hline
    \end{tabular}
  }
\end{table}
\begin{table}[t]
\captionsetup{justification=centering}
  \caption{\label{tab:lmdata} {\it Text data per language in words}}
  \centerline{
    \begin{tabular}{| l | c | c | c | c | c |}
    \hline
    LM set & Sw. & Li. & Bu. & Ta. & So. \\
    \hline \hline
    sup. transcripts & 360K & 400K & 360K & 660K & 430K \\
    unsup. transcripts & 1.4M & 1.2M & 1.5M & 1.5M & 1.5M \\
    Set 1 & 720K & 620K & 720K & 780K & 730K \\
    Set 2 & 77M & 42M & 78M & 120M & 18M \\
    \hline
    \end{tabular}
  }
\end{table}
We run our experiments on five languages used in the IARPA MATERIAL program: Swahili (Sw.), Lithuanian (Li.), Bulgarian (Bu.), Tagalog (Ta.) and Somali (So.).
These corpora vary in size, but all transcribed data is CTS. 
The evaluation sets we use consist entirely of BN data. 
The evaluation set actually consists of two broadcast subsets---news broadcast and ``topical broadcast'', but for the purposes of this work we average the results of these subsets.
The untranscribed sets contains a small amount of CTS data and a large amount of BN data.
Table~\ref{tab:acdata} shows the data in hours per language for the supervised (i.e. transcribed) acoustic data, unsupervised acoustic data, and the evaluation set.

We explore expanding the models with additional LM data, as this has been proven effective for improving hybrid model performance on an unseen domain.
Table~\ref{tab:lmdata} shows the number of words provided by each language model data source, including the supervised acoustic transcripts as well as the unsupervised pseudotranscripts (which are combined with the supervised acoustic transcripts in training baseline semisupervised LMs). 
\textit{Set 1} contains a small amount of parallel data from the MATERIAL program used in training a machine translation system. 
This demonstrates the benefits that even a small amount of additional data can have in adapting the domain.
We have scraped a large amount of text data from the web, as described in \cite{bbn:webdata}, and this data is used in \textit{Set 2}. 
In addition, for Bulgarian and Lithuanian, we include additional data from the Paracrawl corpus \cite{paracrawl} in \textit{Set 2}. 
Using this set in language model training shows the improvements that can be obtained from a much larger---though less refined---corpus of language model data.

\section{Training and Model Details}
\label{sec:models}

\subsection{Hybrid Model}
\label{ssec:hybrid}

Our hybrid systems use a DNN-HMM model implemented in the Kaldi toolkit \cite{jhu:kaldi} and trained using BBN's Sage system \cite{bbn:sage}. 
The input features are 40-dimensional mel-filter cepstrum coefficients (MFCC) features, along with a 100-dimension ivector. 
The HMM structure is the \textit{chain} structure used in \cite{jhu:chain}, which defines a two-state model with a single self-loop. 
The acoustic model generates output at a reduced frame rate of 33Hz, one prediction for every three frames.

The neural network employs a combination of convolution, time-delay, and recurrent layers. 
The basic structure of the network is similar to the Switchboard setup in \cite{cheng2017exploration}, except we prepend an additional 8 convolution layers. 
The MFCC features are transformed back into log mel-filterbank features and passed through the convolutions in parallel with the first three TDNN layers (which process both MFCC and ivector features). 
The output of these subnetworks is concatenated before passing through a ReLU layer and continuing to the alternating LSTM/TDNN portion of the network as described in \cite{cheng2017exploration}. 

The hybrid model is initialized with 1560hrs of multilingual data, the same data used in \cite{bbn:svd}. 
This data is is also used for training the ivector extractor. 
The model is fine-tuned to the target language using the lattice-free MMI objective function \cite{jhu:chain} for one epoch at a constant learning rate of $3 \times 10^{-4}$.
It is then trained with sMBR \cite{karel:smbr} for one epoch at a constant learning rate of $5 \times 10^{-6}$. 
The structure of the model is identical for both the supervised and semisupervised versions.
%

Our hybrid systems use a trigram language model and word-level graphemic pronunciation lexicon.
For the baseline hybrid supervised model, we begin with a dictionary we call \textit{base} that only contains the words in the supervised acoustic transcripts.
Similarly, the lexicon \textit{semisup} contains the words from the semisupervised acoustic transcripts (this varies slightly based on the transcription model).
Finally, we expand the lexicon in \textit{expanded} to include words from a very large web-scraped set (larger than the Set 2 dataset).
The per-language number of words in each lexicon are described in Table~\ref{tab:lex}.

\begin{table}[t]
\captionsetup{justification=centering}
  \caption{\label{tab:lex} {\it Lexicon sizes per language (in words)}}
  \centerline{
    \begin{tabular}{| l | c | c | c | c | c |}
    \hline
    Lexicon & Sw. & Li. & Bu. & Ta. & So. \\
    \hline \hline
    base & 27K & 33K & 22K & 25K & 26K \\
    semisup & 85K & 107K & 72K & 68K & 63K \\
    expanded & 370K & 790K & 750K & 360K & 380K \\
    \hline
    \end{tabular}
  }
\end{table}

\subsection{Sequence-to-sequence Model}
\label{ssec:seq2seq}

Our seq2seq models are trained using the Espresso toolkit \cite{jhu:espresso}. 
The model structure is an encoder-decoder model with attention. For the supervised model, we use a nearly identical setup as \cite{jhu:espresso}. 
The features used are 80-dimensional MFCC features plus 3 additional pitch features. 

The encoder is a CNN-LSTM, with four 2-dimensional convolution layers with (3,3) kernals, with two of the convolution layers downsampling to a total of $\frac{1}{4}$ the time frames. 
The CNN is followed by four bidirectional LSTM layers to produce a 320-dimensional embedding at each downsampled time-step. 
The decoder consists of three unidirectional LSTM layers with Bahdanau attention \cite{bahdanau2016neural}.
All the supervised models use 320 dimension LSTMs, while the semisupervised models vary across languages between 320 and 512 for each LSTM, though the layer sizes are consistent within a given model.
We also add an additional layer in the encoder for the semisupervised models.
The output targets used for all languages are 1000 subword units generated using the SentencePiece implementation of \cite{kudo2018subword}.

The models are trained with cross-entropy. 
The learning rate schedule consists of a 500 step warmup to a constant learning rate of $0.001$ for 150K steps, then decaying linearly over 260K steps to a learning rate of $10^{-5}$, and holding constant until running 180 epochs total and selecting the best model based on validation set WER..
The models are trained with uniform label smoothing with \textit{p} = 0.1 and SpecAugment using the \textit{LD} policy described in \cite{google:specaug}.
The semisupervised training uses the same features, targets, and training procedure.

For seq2seq language models, we train LSTM LMs on the same 1000 subword units for each language.
We use a two-layer, 650-dimensional LSTM for the models trained on acoustic or semisupervised transcripts.
When adding additional LM data, we add a third layer and increase the layer sizes (850-dimension when \textit{Set 1} is added and 1024-dimension when \textit{Set1} and \textit{Set 2} are added).
The model is applied through shallow fusion \cite{bengio:shallowfusion}. 

\subsection{Semisupervised Training}
\label{ssec:sst}

Seed models are trained only on the transcribed data for each language.
These seed models are used to decode the untranscribed data and produce pseudotranscripts.
We treat these pseudotranscripts as though they were ground truth transcripts and combine them with the transcribed data, training the model from scratch to produce our final semisupervised acoustic model.
It can be beneficial to do data selection on the pseudotranscripts \cite{but:semisup, wotherspoon2021improved}, but we keep all data in our experiments..

For the seq2seq seed model, we use the supervised seq2seq baseline \textit{S0} with no shallow fusion.
Based on the results in Table~\ref{tab:sup} (described in more detail in Section~\ref{ssec:sup}), we see an improvement on evaluation WER through shallow fusion.
However, we discovered that using an external LM in pseudotranscription produced instability in model training, with results varying from degradation to training divergence.
As a result, we elected to use no external LM when doing seq2seq pseudotranscription.
For the hybrid seed models, we use two setups.
The first setup is \textit{H1}, where the dictionary is expanded and the LM included the \textit{Set 1} data, while the second setup (\textit{H2}) adds \textit{Set 2} to the LM on top of the \textit{H1} setup.

We use the pseudotranscriptions generated by these seed models combined with the original transcribed audio to retrain the models.
For the seq2seq model, we use the seq2seq pseudotranscripts for retraining.
We train the seq2seq model using the hybrid-generated pseudotranscripts from models \textit{H1} and \textit{H2} as well.
The hybrid system does not suffer from the same domain mismatch problems, and as a result we only examine the pure self-training approach.

\section{Experimental results and analysis}
\label{sec:results}

\begin{table}[t]
  \captionsetup{justification=centering}
  \caption{\label{tab:sup} {\it Supervised results across all five languages on evaluation set}}
  \centerline{
    \begin{tabular}{| c | c | c | c | c | c || c |}
    \hline
    model & Sw. & Li. & Bu. & Ta. & So. & Avg. \\
    \hline \hline
    hybrid bsln (H0) & 46.0 & 49.7 & 42.5 & 47.5 & 61.5 & 49.4 \\
    + lex. expand & 44.5 & 37.2 & 35.3 & 44.6 & 60.3 & 44.4 \\
    + Set 1 LM (H1) & 35.0 & 28.7 & 25.5 & 37.2 & 52.4 & 35.8 \\
    + Set 2 LM (H2) & 31.1 & 24.1 & 20.1 & 32.4 & 49.8 & 31.5 \\
    \hline
    seq2seq bsln (S0) & 60.4 & 63.4 & 58.8 & 68.6 & 82.3 & 66.7 \\
    + external LM & 59.9 & 63.2 & 58.8 & 68.5 & 82.3 & 66.5 \\
    + Set 1 LM (S1) & 58.8 & 61.3 & 56.5 & 67.7 & 82.3 & 65.3 \\
    + Set 2 LM (S2) & 59.0 & 57.3 & 54.1 & 67.2 & 82.2 & 64.0  \\
    \hline
    \end{tabular}
  }
\end{table}

Table~\ref{tab:sup} contains the results for models trained only on supervised acoustic data, and includes the effects of adding additional LM data.
Table~\ref{tab:ssup} contains the results for models trained with semisupervised data using different seed models used for pseudotranscription, again varying the LM data for decoding.
The seed models used are references to models in Table~\ref{tab:sup}.
The rest of this section highlights the different findings of importance.

\subsection{Supervised domain adaptation through LM expansion}
\label{ssec:sup}

From Table~\ref{tab:sup}, we see the differences between the hybrid and seq2seq model configurations.
Even the initial baseline shows a sizable gap between the two systems, with a baseline hybrid model that uses only supervised transcripts having a very large advantage over the baseline seq2seq model---17.3\% WER on average across the five languages.
Clearly domain mismatch is less of a problem for the hybrid model than the seq2seq model, even before any additional improvements are made.

Expanding the LM only widens the gap between the two systems.
Simple lexicon expansion (which cannot be performed on the seq2seq model in its current structure, as we never use a word-level lexicon) yields a 5.0\% WER improvement on average for the hybrid models.
Adding the relatively small (but closely matched) LM data \textit{Set 1} yields an 8.6\% WER improvement, and adding the much larger \textit{Set 2} provides an additional 4.3\% on top of that.
The model overall improves by 17.9\% WER on average simply through the use of additional text information.
For the seq2seq systems, adding an external LM trained on the supervised transcripts provides a very small improvement (0.2\% WER on average).
Expanding it to include additional text data adds only modest improvements---1.2\% WER on average for \textit{Set 1} and 1.3\% WER on top of that for \textit{Set 2} for a total improvement of only 2.7\% WER on average from the addition of a LM trained on our full set of text data.

Coupled with the initial gap between the systems, this puts the best supervised hybrid model at 31.5\% WER on average, while the WER of the best seq2seq model on average is more than double at 64.0\%.
It is clear that for our data condition, hybrid models can be adapted to domain mismatch through the use of external text data.
Seq2seq models cannot be improved to the same degree through text data alone, with improvements from LM only being supplemental. 

\subsection{Seq2seq domain adaptation using hybrid-generated pseudotranscripts}
\label{ssec:seed}


Table~\ref{tab:ssup} shows the variations between using different seed models.
The first result using the \textit{S0} seq2seq model as a seed model shows the true self-learning approach---as we mentioned in Section~\ref{ssec:sst}, using the variants of the seq2seq model with an external LM were unstable, so we elected to use the \textit{S0} model.
On average, true self-learning improves the results by 6.0\% WER on average over the baseline models, with the gain shrinking by 3.3\% WER when considering the baseline with additional LM data added.
Given the operating point, this is a fairly minimal gain.
We attempted to expand this model with an external LM, but the errorful pseudotranscripts caused substantial degradation to the models, so we report no LM expansion on this model.

Seq2seq models can improve to a more competitive operating point when hybrid pseudotranscripts are used.
Using the \textit{H1} model for pseudotranscription yields an average WER of 32.7\% with no LM used, a 28.0\% improvement from the self-learning approach using \textit{S0} for transcription.
This improvement can likely be attributed to the large decrease in errors when using the \textit{H1} model compared to the \textit{S0} model for pseudotranscription, which was made possible by the hybrid models' ability to make strong use of the text data.

We also examine using the \textit{H2} model for pseudotranscription, which used the much larger \textit{Set 2} in LM training.
Note from Table~\ref{tab:sup}, \textit{H2} gives a 4.3\% average WER improvement over \textit{H1}. 
Using the \textit{H2} seed model improves over the \textit{H1}-seeded model by 1.3\% WER on average, showing improved pseudotranscription quality is useful for the seq2seq models.

\subsection{Role of language models in semisupervised performance}
\label{ssec:h2s2s}
\begin{table}[t]
  \captionsetup{justification=centering}
  \caption{\label{tab:ssup} {\it Semisupervised results across all five languages on evaluation set}}
  \centerline{
    \begin{tabular}{| c | c | c | c | c | c | c || c |}
    \hline
    \makecell{seed\\model} & model & Sw. & Li. & Bu. & Ta. & So. & Avg. \\
    \hline\hline
    S0 & s2s SST & 53.2 & 54.5 & 46.5 & 67.4 & 81.8 & 60.7 \\
    \hline\hline
    \multirow{4}{*}{H1} & s2s SST & 29.4 & 25.2 & 22.6 & 35.4 & 50.7 & 32.7 \\
    & + ext LM & 29.2 & 24.9 & 22.4 & 33.7 & 49.7 & 32.0 \\
    & + Set 1 LM & 29.9 & 24.8 & 21.9 & 33.5 & 49.7 & 32.0 \\
    & + Set 2 LM & 28.8 & 23.5 & 20.6 & 33.2 & 49.0 & 31.0 \\
    \hline
    \multirow{3}{*}{H2} & s2s SST & 30.7 & 24.4 & 19.1 & 32.5 & 50.1 & 31.4 \\
    & + ext LM & 27.6 & 24.1 & 18.8 & 30.5 & 47.3 & 29.7 \\
    & + Set 2 LM & 26.9 & 22.2 & 18.7 & 30.1 & 46.9 & 29.0 \\
    \hline\hline
    \multirow{4}{*}{H1} & hyb SST & 30.9 & 26.7 & 22.9 & 34.3 & 48.7 & 32.7 \\
    & + lex. expand & 30.0 & 24.9 & 21.6 & 32.4 & 48.3 & 31.4 \\
    & + Set 1 LM & 29.8 & 24.4 & 20.8 & 32.0 & 48.1 & 31.0 \\
    & + Set 2 LM & 27.8 & 20.8 & 17.3 & 28.5 & 46.6 & 28.2 \\
    \hline
    \multirow{3}{*}{H2} & hyb SST & 29.0 & 24.3 & 20.6 & 29.4 & 46.5 & 30.0 \\
    & + lex. expand & 28.4 & 22.4 & 19.4 & 29.0 & 46.4 & 29.1 \\
    & + Set 2 LM & 27.0 & 20.0 & 16.6 & 27.0 & 45.7 & 27.3 \\
    \hline
    \end{tabular}
  }
\end{table}

We have shown using hybrid-generated pseudotranscripts can be very effective in overcoming the problem of domain mismatch in seq2seq models.
Now that the seq2seq models are producing competitive results, we compare these to the hybrid semisupervised models trained on similar pseudotranscripts.
This shows what sort of gap still remains between the two types of models, and the importance of the language model now that both models are at a similar operating point and have access to reasonable within-domain pseudotranscripts.

In a fair comparison with no additional LM data added (including lexicon expansion for hybrid models), the seq2seq model shows a very slight improvement on average compared to the hybrid counterparts when both use a LM trained only on semisupervised transcripts.
The seq2seq model is 0.7\% WER on average better when \textit{H1} is the seed model, and 0.3\% WER on average better when \textit{H2} is the seed model.
Note the importance of the external LM, it adds on average 0.7\% WER when \textit{H1} is the seed model, and 1.7\% WER when \textit{H2} is the seed model (though much of this is concentrated on Swahili and Tagalog).

However, hybrid models still have significant advantages upon adding additional LM data.
Expanding the dictionary and adding \textit{Set 1} LM data improves the \textit{H1}-seeded hybrid model by 1.7\% WER on average, while the \textit{H1}-seeded seq2seq model gets no improvement from additional LM data.
Overall, the hybrid model is 1.0\% WER better on average when the text data is added.
This gap widens on the \textit{H2}-seeded models.
Lexicon expansion and adding the full LM data provides a 2.7\% WER gain for the hybrid model compared to a 0.7\% average WER improvement for the seq2seq model, leaving the hybrid model 1.7\% better on average when the full LM is included.

Finally, for the \textit{H1}-seeded models, we can add the \textit{Set 2} LM data to observe the effect of additional text data not seen in pseudotranscription.
The hybrid model gets an improvement of 2.8\% WER on average compared to the 1.0\% WER gain from the seq2seq model, again showing the power that additional LM data can provide hybrid models that our current seq2seq models are unable to replicate.
Perhaps other approaches beyond shallow fusion like deep \cite{gulcehre2015using} and cold fusion \cite{sriram2017cold} would provide further gains from the additional LM data, but previous work has seen limited gains from these techniques \cite{inaguma2019transfer, toshniwal2018comparison}.

\section{Conclusion}
\label{sec:conclusion}

In this work, we have shown that seq2seq ASR systems struggle in the face of a domain mismatch between training and test in a low resource environment.
This domain mismatch is not as much of a problem for hybrid systems, which are more generalizable and less data hungry.
Additionally, hybrid systems can easily be adapted to a new domain through dictionary and language model expansion.
While applying external language models does provide a small gain to seq2seq models, it is not able to overcome a dramatic domain shift.

To this end, we used semisupervised training to generate transcribed acoustic data from an untranscribed set of target matched data using the self-learning paradigm.
Using a seq2seq model for pseudotranscription provides a small gain, improving on average from 66.7\% WER to 60.7\% compared to the baseline seq2seq model. 
However, using a hybrid model for pseudotranscription yields a much larger gain, improving from 66.7\% WER to 29.0\% WER when using all of the LM data available.
This operating point is much closer to competitive systems.

We hypothesize that using the hybrid-generated pseudotranscripts for seq2seq models could be thought of as an efficient form of knowledge transfer.
It is clear from our results that seq2seq models are unable to use external LMs to dramatically alter model predictions as is necessary for domain mismatch.
Seq2seq models seem to use external LMs for smaller, supplemental shifts.
Hybrid generated pseudotranscripts inject this LM information directly into the seq2seq model in a way that allows the model to make better use of the information to more substantially alter predictions.

A gap still remains between semisupervised hybrid and seq2seq models, primarily due to the hybrid models' ability to take advantage of LM data outside of the pseudotranscripts.
Despite this, we are encouraged by these results.
Improving the seq2seq models from being in some cases completely unusable to being competitive with well-trained hybrid systems is a strong step forward.
Current ASR research has heavily shifted focus towards seq2seq models, and we are now able to apply these models to tasks where previously they would have been unusable without domain-matched transcribed data.

%
For future work, it would be interesting to investigate a data condition with a similar mismatch where significantly more out-of-domain transcribed data was available.
It would also be interesting to investigate domains that are not easily captured through text. 
Our current seq2seq models are fairly simple, so we will investigate more powerful model architectures such as the Conformer \cite{google:conformer}, as well as additional types of external LMs for seq2seq models.
We also plan to explore unsupervised pretraining \cite{baevski2020wav2vec}, especially as it has been shown to be complementary to self-learning \cite{xu2020self}.

\section{Acknowledgements}
\label{sec:ack}

This work was supported by the Intelligence Advanced Research Projects Activity (IARPA) via Department of Defense US Air Force Research Laboratory contract number FA8650-17-C-9118.



\vfill\pagebreak

\bibliographystyle{IEEEbib}
\bibliography{e2e-sst}

\begin{thebibliography}{10}

\bibitem{ibm:swbd300}
Zoltan Tuske, George Saon, Kartik Audhkhasi, and Brian Kingsbury,
\newblock ``Single headed attention based sequence-to-sequence model for
  state-of-the-art results on switchboard-300,''
\newblock in {\em Proceedings of INTERSPEECH}, 2020.

\bibitem{aachen:e2e}
Albert Zeyer, Kazuki Irie, Ralf Schl{\"{u}}ter, and Hermann Ney,
\newblock ``Improved training of end-to-end attention models for speech
  recognition,''
\newblock {\em CoRR}, vol. abs/1805.03294, 2018.

\bibitem{google:conformer}
Anmol Gulati, James Qin, Chung-Cheng Chiu, et~al.,
\newblock ``Conformer: Convolution-augmented transformer for speech
  recognition,''
\newblock in {\em Proceedings of INTERSPEECH}, 2020.

\bibitem{google:specaug}
Daniel~S. Park, William Chan, Yu~Zhang, Chung-Cheng Chiu, Barret Zoph, Ekin~D.
  Cubuk, and Quoc~V. Le,
\newblock ``Specaugment: A simple data augmentation method for automatic speech
  recognition,''
\newblock in {\em Proceedings of INTERSPEECH}, 2019.

\bibitem{ibm:e2ekws}
A.~{Rosenberg}, K.~{Audhkhasi}, A.~{Sethy}, B.~{Ramabhadran}, and M.~{Picheny},
\newblock ``End-to-end speech recognition and keyword search on low-resource
  languages,''
\newblock in {\em Proceedings of ICASSP}, 2017, pp. 5280--5284.

\bibitem{cho2018multilingual}
Jaejin Cho, Murali~Karthick Baskar, Ruizhi Li, Matthew Wiesner, Sri~Harish
  Mallidi, Nelson Yalta, Martin Karafiat, Shinji Watanabe, and Takaaki Hori,
\newblock ``Multilingual sequence-to-sequence speech recognition: architecture,
  transfer learning, and language modeling,''
\newblock in {\em Proceedings of SLT}. IEEE, 2018, pp. 521--527.

\bibitem{wang2020improving}
Changhan Wang, Juan Pino, and Jiatao Gu,
\newblock ``Improving cross-lingual transfer learning for end-to-end speech
  recognition with speech translation,''
\newblock {\em arXiv preprint arXiv:2006.05474}, 2020.

\bibitem{karafidt2018analysis}
Martin Kar{\'a}fidt, Murali~Karthick Baskar, Karel Vesel{\`y}, Franti{\v{s}}ek
  Gr{\'e}zl, Luk{\'a}{\v{s}} Burget, and Jan {\v{C}}ernock{\`y},
\newblock ``Analysis of multilingual blstm acoustic model on low and high
  resource languages,''
\newblock in {\em Proceedings of ICASSP}. IEEE, 2018, pp. 5789--5793.

\bibitem{shetty2020improving}
Vishwas~M Shetty, Metilda Sagaya~Mary NJ, and S~Umesh,
\newblock ``Improving the performance of transformer based low resource speech
  recognition for indian languages,''
\newblock in {\em Proceedings of ICASSP}. IEEE, 2020, pp. 8279--8283.

\bibitem{srivastava2018interspeech}
Brij Mohan~Lal Srivastava, Sunayana Sitaram, Rupesh~Kumar Mehta, Krishna~Doss
  Mohan, Pallavi Matani, Sandeepkumar Satpal, Kalika Bali, Radhakrishnan
  Srikanth, and Niranjan Nayak,
\newblock ``Interspeech 2018 low resource automatic speech recognition
  challenge for indian languages.,''
\newblock in {\em SLTU}, 2018, pp. 11--14.

\bibitem{inaguma2019transfer}
Hirofumi Inaguma, Jaejin Cho, Murali~Karthick Baskar, Tatsuya Kawahara, and
  Shinji Watanabe,
\newblock ``Transfer learning of language-independent end-to-end asr with
  language model fusion,''
\newblock in {\em Proceedings of ICASSP}. IEEE, 2019, pp. 6096--6100.

\bibitem{but:semisup}
Karel Vesely, Mirko Hannemann, and Lukas Burget,
\newblock ``Semi-supervised training of deep neural networks,''
\newblock in {\em Proceedings of ASRU}, 2013.

\bibitem{bbn:semisup}
Scott Novotney, Richard Schwartz, and Jeff Ma,
\newblock ``Unsupervised acoustic and language model training with small
  amounts of labelled data,''
\newblock in {\em Proceedings of ICASSP}, 2009.

\bibitem{ms:semisup}
Yan Huang, Dong Yu, Yifan Gong, and Chaojun Liu,
\newblock ``Semi-supervised gmm and dnn acoustic model training with
  multi-system combination and confidence re-calibration,''
\newblock in {\em Proceedings of INTERSPEECH}, 2013.

\bibitem{wotherspoon2021improved}
Shannon Wotherspoon, William Hartmann, Matthew Snover, and Owen Kimball,
\newblock ``Improved data selection for domain adaptation in asr,''
\newblock in {\em Proceedings of ICASSP}, 2021.

\bibitem{Kahn_2020}
Jacob Kahn, Ann Lee, and Awni Hannun,
\newblock ``Self-training for end-to-end speech recognition,''
\newblock in {\em Proceedings of ICASSP}, 2020.

\bibitem{chen2020semisupervised}
Yang Chen, Weiran Wang, and Chao Wang,
\newblock ``Semi-supervised asr by end-to-end self-training,''
\newblock in {\em Proceedings of INTERSPEECH}, 2020.

\bibitem{park2020improved}
Daniel~S. Park, Yu~Zhang, Ye~Jia, Wei Han, Chung-Cheng Chiu, Bo~Li, Yonghui Wu,
  and Quoc~V. Le,
\newblock ``Improved noisy student training for automatic speech recognition,''
  2020.

\bibitem{bbn:webdata}
Le~Zhang, Damianos Karakos, William Hartmann, Roger Hsiao, Richard Schwartz,
  and Stavros Tsakalidis,
\newblock ``Enhancing low resource keyword spotting with automatically
  retrieved web documents,''
\newblock in {\em Proceedings of INTERSPEECH}, 2015.

\bibitem{paracrawl}
Miquel Espl{\`a}, Mikel Forcada, Gema Ram{\'\i}rez-S{\'a}nchez, and Hieu Hoang,
\newblock ``{P}ara{C}rawl: Web-scale parallel corpora for the languages of the
  {EU},''
\newblock in {\em Proceedings of Machine Translation Summit XVII Volume 2:
  Translator, Project and User Tracks}, Dublin, Ireland, Aug. 2019, pp.
  118--119, European Association for Machine Translation.

\bibitem{jhu:kaldi}
Daniel Povey, Arnab Ghoshal, Gilles Boulianne, et~al.,
\newblock ``The {Kaldi} speech recognition toolkit,''
\newblock in {\em Proceedings of ASRU}, 2011.

\bibitem{bbn:sage}
Roger Hsiao, Ralf Meermeier, Tim Ng, et~al.,
\newblock ``Sage: The new {BBN} speech processing platform,''
\newblock in {\em Proceedings of INTERSPEECH}, 2016.

\bibitem{jhu:chain}
Daniel Povey, Vijayaditya Peddinti, Daniel Galvez, et~al.,
\newblock ``Purely sequence-trained neural networks for {ASR} based on
  lattice-free {MMI},''
\newblock in {\em Proceedings of INTERSPEECH}, 2016.

\bibitem{cheng2017exploration}
Gaofeng Cheng, Vijayaditya Peddinti, Daniel Povey, Vimal Manohar, Sanjeev
  Khudanpur, and Yonghong Yan,
\newblock ``An exploration of dropout with lstms.,''
\newblock in {\em Proceedings of INTERSPEECH}, 2017, pp. 1586--1590.

\bibitem{bbn:svd}
Francis Keith, William Hartmann, Man{-}Hung Siu, Jeff~Z. Ma, and Owen Kimball,
\newblock ``Optimizing multilingual knowledge transfer for time-delay neural
  networks with low-rank factorization,''
\newblock in {\em Proceedings of ICASSP}. 2018, pp. 4924--4928, {IEEE}.

\bibitem{karel:smbr}
Karel Vesel{\`y}, Arnab Ghoshal, Luk{\'a}s Burget, and Daniel Povey,
\newblock ``Sequence-discriminative training of deep neural networks,''
\newblock in {\em Proceedings of INTERSPEECH}, 2013.

\bibitem{jhu:espresso}
Yiming Wang, Tongfei Chen, Hainan Xu, Shuoyang Ding, Hang Lv, Yiwen Shao,
  Nanyun Peng, Lei Xie, Shinji Watanabe, and Sanjeev Khudanpur,
\newblock ``Espresso: A fast end-to-end neural speech recognition toolkit,''
\newblock in {\em Proceedings of ASRU}, 2019.

\bibitem{bahdanau2016neural}
Dzmitry Bahdanau, Kyunghyun Cho, and Yoshua Bengio,
\newblock ``Neural machine translation by jointly learning to align and
  translate,''
\newblock in {\em Proceedings of ICLR}, 2015.

\bibitem{kudo2018subword}
Taku Kudo,
\newblock ``Subword regularization: Improving neural network translation models
  with multiple subword candidates,''
\newblock in {\em Proceedings of ACL}, 2018.

\bibitem{bengio:shallowfusion}
{\c{C}}aglar G{\"{u}}l{\c{c}}ehre, Orhan Firat, Kelvin Xu, Kyunghyun Cho,
  Lo{\"{\i}}c Barrault, Huei{-}Chi Lin, Fethi Bougares, Holger Schwenk, and
  Yoshua Bengio,
\newblock ``On using monolingual corpora in neural machine translation,''
\newblock {\em CoRR}, vol. abs/1503.03535, 2015.

\bibitem{gulcehre2015using}
Caglar Gulcehre, Orhan Firat, Kelvin Xu, Kyunghyun Cho, Loic Barrault, Huei-Chi
  Lin, Fethi Bougares, Holger Schwenk, and Yoshua Bengio,
\newblock ``On using monolingual corpora in neural machine translation,''
\newblock {\em arXiv preprint arXiv:1503.03535}, 2015.

\bibitem{sriram2017cold}
Anuroop Sriram, Heewoo Jun, Sanjeev Satheesh, and Adam Coates,
\newblock ``Cold fusion: Training seq2seq models together with language
  models,''
\newblock {\em arXiv preprint arXiv:1708.06426}, 2017.

\bibitem{toshniwal2018comparison}
Shubham Toshniwal, Anjuli Kannan, Chung-Cheng Chiu, Yonghui Wu, Tara~N Sainath,
  and Karen Livescu,
\newblock ``A comparison of techniques for language model integration in
  encoder-decoder speech recognition,''
\newblock in {\em Proceedings of SLT}. IEEE, 2018, pp. 369--375.

\bibitem{baevski2020wav2vec}
Alexei Baevski, Henry Zhou, Abdelrahman Mohamed, and Michael Auli,
\newblock ``wav2vec 2.0: A framework for self-supervised learning of speech
  representations,'' 2020.

\bibitem{xu2020self}
Qiantong Xu, Alexei Baevski, Tatiana Likhomanenko, Paden Tomasello, Alexis
  Conneau, Ronan Collobert, Gabriel Synnaeve, and Michael Auli,
\newblock ``Self-training and pre-training are complementary for speech
  recognition,''
\newblock {\em arXiv preprint arXiv:2010.11430}, 2020.

\end{thebibliography}

\end{document}